\documentclass{bvm} 

\makeatletter
\let\blx@rerun@biber\relax
\makeatother

\begin{document}

\newcommand{\bvmyear}{2025}

\selectlanguage{english} 

\title{A Systematic Analysis of Input Modalities for Fracture Classification of the Paediatric Wrist}


\titlerunning{Analysis of Input Modalities for Fracture Classification}

\author{Ron Keuth\inst{1},
    Maren Balks\inst{3},
    Sebastian Tschauner\inst{4},
    Ludger Tüshaus\inst{2},
    Mattias Heinrich\inst{1}
}

\authorrunning{Keuth et al.}

\institute{
\inst{1} Institute of Medical Informatics, University of Lübeck \\
\inst{2} Paediatric Surgery, University Hospital Schleswig-Holstein\\
\inst{3} Institut of Radiology and Nuclear Medicine, University Hospital Schleswig-Holstein\\
\inst{4} Paediatric Radiology, Medical University of Graz}

\email{r.keuth@uni-luebeck.de}

\maketitle

\begin{abstract}
Fractures, particularly in the distal forearm, are among the most common injuries in children and adolescents, with approximately 800\,000 cases treated annually in Germany. The AO/OTA system provides a structured fracture type classification, which serves as the foundation for treatment decisions. Although accurately classifying fractures can be challenging, current deep learning models have demonstrated performance comparable to that of experienced radiologists.
While most existing approaches rely solely on radiographs, the potential impact of incorporating other additional modalities, such as automatic bone segmentation, fracture location, and radiology reports, remains underexplored. In this work, we systematically analyse the contribution of these three additional information types, finding that combining them with radiographs increases the AUROC from 91.71 to 93.25. Our code is available on \href{https://github.com/multimodallearning/AO_Classification}{GitHub}.
\end{abstract}

\section{Introduction}
Fractures, particularly in the distal forearm, are among the most common injuries in children and adolescents, with approximately 800\,000 cases treated annually in Germany. The risk of suffering a fracture by the end of the growing years is estimated to be between 15\ts\% and 45 \ts\% \cite{Krause2005}.
However, due to child-specific factors like skeletal growth, fracture treatment decisions can differ fundamentally from those in adults.
To determine the appropriate treatment, factors beyond the patient’s age and growth stage must be considered, including fracture location, degree of displacement, fracture pattern, and any accompanying injuries.
These considerations are incorporated in the AO/OTA Paediatric Comprehensive Classification of Long Bone Fractures system, developed and validated specifically for paediatric fractures \cite{AOSystem2006}.
Since 2006, it is now the most widely adopted framework worldwide, serving as the standard for documentation, communication, and treatment planning for paediatric fractures.\par
In recent years, deep learning has supported clinicians in radiograph interpretation, matching the performance of human experts in fracture detection \cite{Kuo2022SurveyFractureDetection}.
For fracture classification, evidence shows similar results for distal forearm fractures (AUC of 0.89)\cite{Multiclass_deep_learning_based_AO}.
While further work focuses on tweaking the feature extraction by comparing different CNN-based encoders (AUC of 0.94)\cite{DeepLearningModelForDRF}, others employ a classical and deep learning-based feature fusion (F1 of 0.81)\cite{AO_combine_classic_dl_feat}.
Hierarchical classification approaches, including multilabel subgrouping for ankle fractures (AUC of 0.9)\cite{ankle_fracture_classification} and multistage models for radius fractures (AUC of 0.82)\cite{TwoStageEnsemble}, offer further refinement.
However, while some multistage methods use fracture location for classification\cite{TwoStageEnsemble}, the impact of additional modalities remains underexplored.
Further motivation is provided by other classification settings, demonstrating the positive impact of intermediate representations, e.g., depth, and flow field in autonomous driving \cite{Zhou2019}, landmarks for facial emotion classification \cite{Hasani2017}, as well as segmentation facial expression classification \cite{meine_BA}.
In this work, we systematically analyse the contribution of three modalities, namely automatic bone segmentation, fracture location, and radiology reports as additional input to the radiograph for fracture classification of the paediatric wrist within the AO/OTA system.
We found providing all modalities improves the performance, with the impact of fracture location being significant.

\section{Materials and Methods}
\subsection{Dataset}
The GRAZPEDWRI-DX dataset\cite{nagy_pediatric_2022} is a publicly available dataset containing over 20k radiographs of 5900 children and adolescents.
For fractures, it provides bounding box annotations, their AO/OTA codes, and the corresponding radiology report for each radiograph.
Since we only have segmentations for the 10k AP perspectives, we exclude the lateral views.
For our classification task, we use the eight most common AO/OTA classes (adding "no fracture") and split the dataset into 7809/1978 radiographs for training/testing.

\begin{figure}
    \centering
    \includegraphics[width=\textwidth]{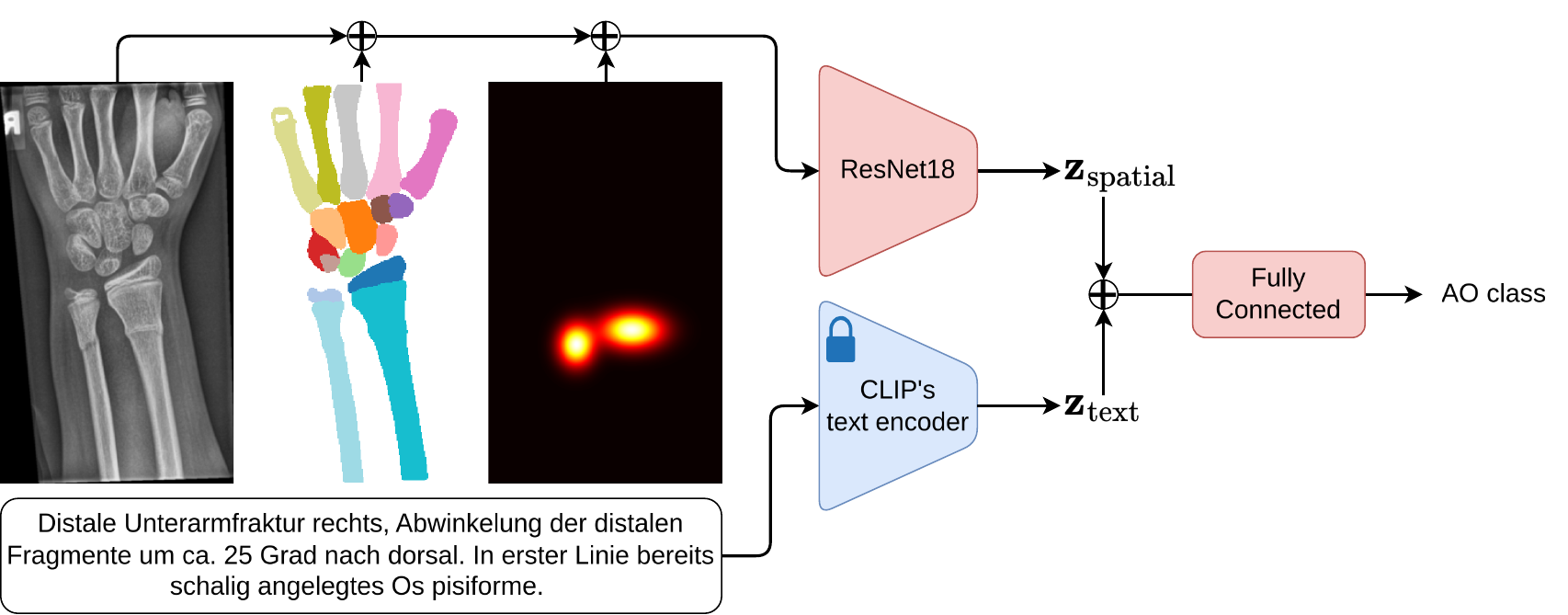}
    \caption{Usage of the four modalities for fracture classification. The radiograph, the bone segmentation, and the heatmap encoding the fractures' location are feed to the ResNet18. The report embedding of the frozen CLIP text encoder $\mathbf{z}_\text{text}\in\mathbb{R}^{512}$ is fused to the ResNet's latent vector $\mathbf{z}_\text{spatial}\in\mathbb{R}^{512}$. $\bigoplus$ describes channel-wise concatenation.}
    \label{fig:method_overview}
\end{figure}

\subsection{Setup}
Fig. \ref{fig:method_overview} shows a schematic overview, how we incorporate the additional modalities into our classification pipeline.
For the spatial input (radiograph, bone segmentation, and fracture location), we concatenate them channelwise $\bigoplus$ and feet them into a ResNet18 embedding them into $\mathbf{z}_\text{spatial}\in\mathbb{R}^{512}$.
The non-learnable radiology report embedding $\mathbf{z}_\text{text}\in\mathbb{R}^{512}$ is then concatenated to $\mathbf{z}_\text{spatial}$, constructing the latent space for the classifier, which linear projects its 1024 dimensions onto the eight fracture classes likelihoods.

\subsubsection{Different modalities}
In this section, we will briefly introduce the four different modalities used by our classifier.
1) The \emph{radiograph} itself, which we z normalize and resize to $384\times 224$, where small details like fractures are still visible.
2) The multilabel \emph{segmentations} of 17 bones. Those were created in a semi-supervised setting, starting with 40 annotated images and using label propagation with a SAM-based mask refinement to enable training on all unlabelled data (reaching a Dice of 84.2\ts\% on nine annotated test images)\cite{keuth2024samcarriesburdensemisupervised}.
3) We encode the \emph{fracture's locations} $(h_f,w_f)$ using a heatmap $\mathbf{H}\in\mathbb{R}^{H\times W}$.
For this, we extract the parameters of the multivariate Gaussian kernel from the fracture's bounding box, where its centre $(h_c,w_c)$ represents the means and its axis lengths $(l_h, l_w)$ the standard derivations:
\begin{equation}
    \mathbf{H}(h,w) = \frac{1}{2\pi l_h l_w}\exp\left[-\frac{1}{2}\left(\frac{(h-h_f)^2}{l_h^2}+\frac{(w-w_f)^2}{l_w^2}\right)\right].
\end{equation}
We fancy more peaky Gaussian blobs with $l_h/2$ and $l_w/2$ and normalize $\mathbf{H}$ to [0,1].
While we use the ground truth bounding boxes during training, we use a YOLOv10x to estimate the fracture locations during inference.
We train the model on the same train/test split and achieve an accuracy of 90\ts\% and an area under the \mbox{precision-recall} curve of 93.4\ts\% on the test data.
4) As a last modality, we include the \emph{radiology report}, which we encode with a distilled variant of a pretrained multilingual BERT model (\texttt{distilbert-base-cased} on Hugging Face).
Since an end-to-end approach, trying to classify solely based on the text embedding, did not converge (15\ts\% precision, 100\ts\% recall), we follow a more complex CLIP approach, where the text embedding gets aligned with the corresponding radiograph embedding in a self-supervised fashion.
The radiograph embedding is thereby learned by a ResNet18 encoder.
We employ a \mbox{two-layer} MLP on top of the BERT model, projecting the text embedding to 512 dimensions.
During training (same routine as in Sec. \ref{sec.training_routine}, batch size of 256), we minimize the InfoNCE loss, forcing CLIP to lean semantically cluster and only adapt the ResNet18 and projection MLP, leaving the text model frozen.
Since the usability of the learned clusters for our fracture classification is not guaranteed, we provide a proof of concept, where we fit a single fully-connected layer on CLIP's frozen latent space.
As a comparison, we choose ImageNet features (provided by a ResNet18) and the latent representation of an Autoencoder, which we train with an L1-norm to reconstruct the radiographs (same routine as in Sec. \ref{sec.training_routine} with no data augmentation).

\subsubsection{Training routine}\label{sec.training_routine}
Since a wrist can suffer from multiple fractures, we formulate our problem as a multilabel classification.
Hence, use the binary cross entropy, which is weighted to balance recall and precision.
The loss is minimized by the ADAM optimiser with a learning rate of $1e-3$ over 100 epochs and a batch size of 64.
To prevent overfitting, we introduce dropout with a rate of 60\ts\% on $[\mathbf{z}\_\text{spatial}, \mathbf{z}\_\text{text}]^\intercal$ and use spatial data augmentation by applying affine transformation, where the parameters are random sampled from $\mathcal{U}(-30^\circ, 30^\circ)$ for rotation, from $\mathcal{U}(-10\ts\%, 10\ts\%)$ for translation, and $\mathcal{U}(85\ts\%, 115\ts\%)$ for scaling (percentage of spatial image dimensions).
Our code is available on \href{https://github.com/multimodallearning/AO_Classification}{GitHub}\footnote{\url{https://github.com/multimodallearning/AO_Classification}}.

\section{Results}
We train a model for each of the eight possible modality combinations of radiograph, multilabel bone segmentation, fracture's location (as heatmap), and CLIP's text embedding of the radiology report.
Tab. \ref{tab:quantitative} shows a selection of the quantitative results represented by accuracy, F1, precision, recall, and area under the receiver operating characteristic curve (AUROC), which are calculated independently for each class and averaged in a macro-fashion.
The standard approach, only considering the radiograph, already yields a competitive baseline with an AUROC of 91.71\ts\% and F1 of 60.69\ts\%.
While providing the segmentation as additional input, the balance of precision and recall shifted towards the latter by remaining a comparable AUROC but decreasing F1.
However, using the fracture's heatmap instead of segmentation, we obtain an increase of recall by over 5\ts\% compared to the baseline, resulting in a rise of 1.3\ts\% in AUROC.
When both are used, we obtain the best trade-off between precision and recall, yielding the overall best F1 of 63.51\ts\%.
CLIP's report embedding further increases recall and AUROC to 88.83\ts\% and 93.26\ts\%, respectively.\par
To create further insights into the individual improvement each modality provides, we compare all experiments with the equivalent experiments without this modality.
We test for significance with the Wilcoxon paired signed-rank test on the AUROCs of the eight classes with a significance level $\alpha=0.05$.
Since the test includes four comparisons for each modality, we reduce $\alpha'=\frac{\alpha}{4}=0.0125$ for multiple comparisons with the Bonferroni correction.
In our setting with $\alpha'$, we could not find any significant improvement in providing the bone segmentation ($0.5<p_{\max}$), fracture heatmap ($0.019<p_{\max}$), or CLIP's report embedding ($0.07<p_{\max}$) independently of the availability of the other modalities.
However, when just the addition to the radiograph is considered, the fracture heatmap provides a significant improvement ($p<0.007<\frac{\alpha}{1}$).

\begin{table}[]
    \centering
    \caption{Quantitative results on the test split. \checkmark describes the use of the modality radiograph (Img), bone segmentation (BoneSeg), heatmap encoding fracture's location (FracLoc), and CLIP's text embedding of the radiology report (Report). All metrics are computed over the eight fracture types independently and macro averaged.}
    \begin{tabular*}{\textwidth}{c@{\extracolsep\fill}cccccccccc}
        \hline
        Img & BoneSeg & FracLoc & Report & Accuracy & F1 & Precision & Recall & AUROC \\
        \hline
        \checkmark &  &  &  & 85.16 & 60.69 & 49.25 & 82.12 & 91.71 \\
        \checkmark & \checkmark &  &  & 84.16 & 59.26 & 47.21 & 85.09 & 91.60 \\
        \checkmark &  & \checkmark &  & 84.90 & 61.66 & 48.75 & 87.40 & 93.08 \\
        \checkmark & \checkmark & \checkmark &  & \textbf{86.89} & \textbf{63.51} & \textbf{53.67} & 84.04 & 93.15 \\
        \checkmark & \checkmark & \checkmark & \checkmark & 84.80 & 61.70 & 49.51 & \textbf{88.83} & 93.16 \\
        \checkmark &  & \checkmark & \checkmark & 86.60 & 63.47 & 51.73 & 84.48 & \textbf{93.26} \\
        \hline
    \end{tabular*}
\label{tab:quantitative}
\end{table}

Regarding the linear evaluation of CLIP's latent space as proof of concept concerning its usability for fracture classification, Tab. \ref{tab:ablation_clip_lin_eval} shows, that both CLIP image and text encoder provide suitable features, yielding an AUROC of 85.4 and 83.77 respectively.
With this, CLIP outperforms the features provided by an ImageNet pretrained ResNet18 and an autoencoder trained in reconstructing the radiographs.
Since we are interested in using the reports as an additional modality, we only considered CLIP's text embeddings.

\begin{figure}[b]
    \setlength{\figwidth}{.40\textwidth}
    \centering
    \begin{subfigure}[t]{\figwidth}
        \centering
		\includegraphics[width=\textwidth]{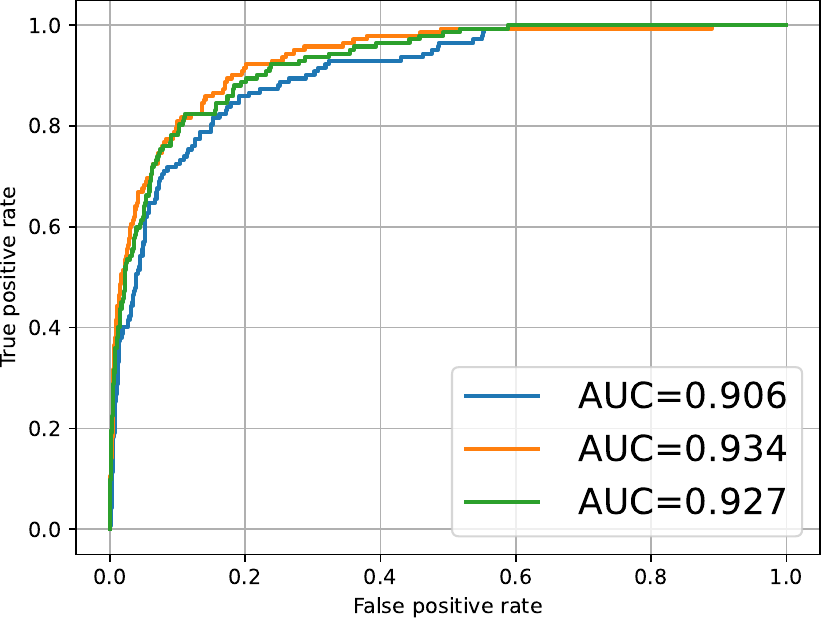}
		\caption{23-M/2.1}
	\end{subfigure}
    \hspace{.05\figwidth}
    \begin{subfigure}[t]{\figwidth}
        \centering
		\includegraphics[width=\textwidth]{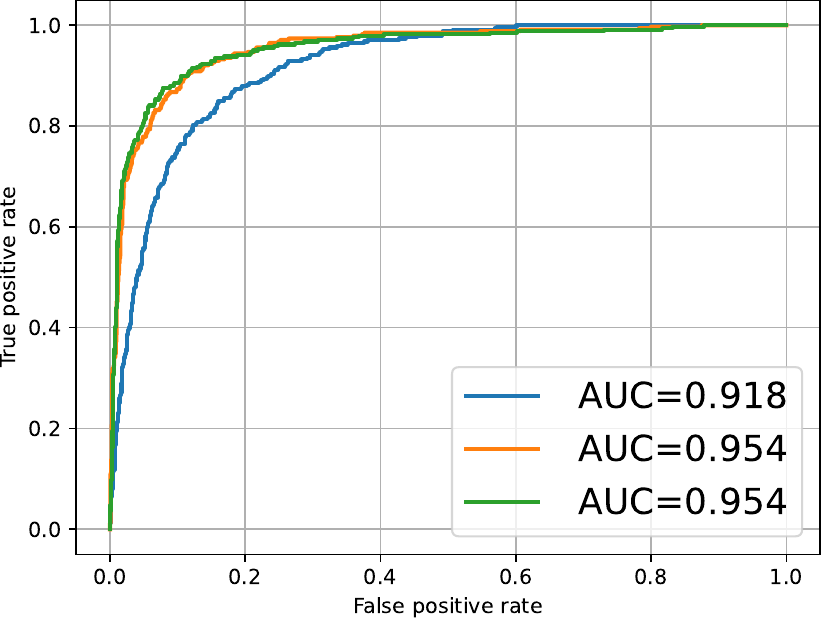}
		\caption{23u-E/7}
	\end{subfigure}
    \caption{AUROC comparison for fracture types (AO/OTA codes in captions) shows the benefit of including fracture location as input. Blue: Img, orange: Img + FracLoc, green: all modalities.}
    \label{fig:roc_plots}
\end{figure}

Fig. \ref{fig:roc_plots} shows the ROC of two fracture types, where the benefit of using multiple modalities as input is especially visible.
Both AO/OTA classes correspond to torus fracture (incomplete fracture with intact periost) of the radius and ulna.
Due to their subtle and incomplete fracture lines, torus fractures can be especially hard to notice in radiographs.
In such cases, the fracture heatmap can provide guidance.

\begin{SCtable}[]
    \centering
    \caption{Linear evaluation of CLIP's latent space demonstrates its potential for fracture classification, compared to ImageNet features and an autoencoder's latent vector. All encoders are frozen.}
    \begin{tabular}{lccc}
        \hline
        Encoder & Accuracy & F1 & AUROC \\
        \hline
        CLIP's img encoder & 73.97 & 48.10 & 85.40 \\
        CLIP's txt encoder & 72.35 & 46.90 & 83.77 \\
        ResNet18 ImageNet & 67.80 & 40.68 & 77.49 \\
        AE on GRAZPEDWRI & 64.65 & 39.23 & 76.33 \\
        \hline
        \end{tabular}
    \label{tab:ablation_clip_lin_eval}
\end{SCtable}

\section{Discussion}
Our analysis reveals that providing bone segmentation, fracture location, and radiology reports in addition to the radiographs increases the performance for fracture classification regarding accuracy, precision, recall, and AUROC.
However, by adding each modality alone to the radiograph, we could only prove the significance of the fracture location.
The added value by the fracture's location becomes visible by considering the ROCs of two AO/OTA codes (Fig. \ref{fig:roc_plots}) describing a torus fracture with fine fracture lines, which can be easily missed in the radiograph alone or is not represented in the bone segmentation.
However, regarding the results, including bone segmentations, it should be noted, that since the GRAZPEDWRI dataset natively lacks segmentation, the use of our predicted ones (Dice of 84.2\ts\% on nine test samples) still has some potential for improvement.
Hence, those results could be underestimated due to their quality.
While we demonstrated that CLIP's self-supervised training enables the use of radiology reports as an input modality for our fracture classification (Tab. \ref{tab:ablation_clip_lin_eval}), it alone does not provide a significant improvement. 
This could be due to the relatively short and not necessary detailed description of the fractures in the radiology report, lacking direct information on the AO/OTA class (Fig. \ref{fig:method_overview}).
In future work, we will investigate the different modalities' potential for reducing the amount of training data  while remaining the same performance, as it has been already demonstrated for segmentation in other settings \cite{meine_BA}.
Moreover, considering a hierarchical \cite{ankle_fracture_classification} or multistage \cite{TwoStageEnsemble} approach instead of a multilabel one could be beneficial, as both are better suited to the hierarchy of the AO/OTA system.

\acknowledgement
This research has been funded by the state of Schleswig-Holstein, Grant Number 22023005.

\printbibliography

\end{document}